\documentclass[runningheads]{llncs}
\usepackage{graphicx}
\usepackage{booktabs}
\usepackage{makecell}
\usepackage{mathtools}
\begin{document}

\title{Detect Depression from Social Networks with Sentiment Knowledge Sharing}
\author{Yan Shi\inst{1} \and
Yao Tian\inst{1} \and Chengwei Tong\inst{1} \and Chunyan Zhu\inst{2, 3} \and
Qianqian Li\inst{3} \and Mengzhu Zhang\inst{2} \and Wei Zhao\inst{4} \and Yong Liao\inst{1} \and Pengyuan Zhou\inst{1*}}
\institute{University of Science and Technology of China, Hefei, China \\
\email{\{syan0724, tyzkd, cwtong\}@mail.ustc.edu.cn, \{yliao, pyzhou\}@ustc.edu.cn} 
\and Anhui Medical University, Hefei, China\\
\and the Second Affiliated Hospital of Anhui Medical University, Hefei, China\\
\email{ayswallow@126.com, \{lqqian187, zhangmengzhu2021\}@163.com}
\and Anhui ERCIASII, Anhui University of Technology, Ma’anshan, China\\
\email{zhaowei@ahut.edu.cn} }
\maketitle              % typeset the header of the contribution
\begin{abstract}
%抑郁检测的重要性
Social network plays an important role in propagating people's viewpoints, emotions, thoughts, and fears. Notably, following lockdown periods during the COVID-19 pandemic, the issue of depression has garnered increasing attention, with a significant portion of individuals resorting to social networks as an outlet for expressing emotions. Using deep learning techniques to discern potential signs of depression from social network messages facilitates the early identification of mental health conditions. 
Current efforts in detecting depression through social networks typically rely solely on analyzing the textual content, overlooking other potential information. In this work, we conduct a thorough investigation that unveils a strong correlation between depression and negative emotional states. The integration of such associations as external knowledge can provide valuable insights for detecting depression. Accordingly, we propose a multi-task training framework, DeSK, which utilizes shared sentiment knowledge to enhance the efficacy of depression detection. Experiments conducted on both Chinese and English datasets demonstrate the cross-lingual effectiveness of DeSK.

\end{abstract}
\section{Introduction}
In recent years, there has been a growing focus on the subject of mental health, drawing significant attention from the general public. Particularly following the outbreak of the COVID-19 pandemic, a surge in the prevalence of common mental health disorders has been observed~\cite{chandola2022mental}. And among these disorders, depression stands out as the most prevalent form, exhibiting a strong correlation with substantial morbidity and mortality rates~\cite{cuijpers2020treatment}. Traditional methods for depression diagnosis usually use interviews with patients or self-report questionnaires, which are time-consuming and error-prone.

The social network offers a pathway for capturing pertinent behavioral attributes pertaining to an individual's cognition, emotional state, communication patterns, daily activities, and social interactions. The emotions conveyed and the linguistic patterns employed in posts on social networks can potentially serve as indicators of prevailing sentiments such as feelings of insignificance, culpability, powerlessness, and intense self-disdain, which are characteristic of major depressive disorder~\cite{de2013predicting}. Hence, it becomes paramount to comprehend and analyze the emotions that individuals convey through social networks, especially during difficult times like the pandemic. Furthermore, the timely identification of initial indicators of depression is of great importance, as it enables prompt intervention and assistance for those in need.

There is a large amount of existing work analyzing depression using social network data. Early works are usually based on statistical or traditional machine-learning approaches. \cite{shen2017depression} proposed a multi-modal depressive dictionary learning model specifically for detecting users with depressive tendencies on Twitter. Recently, deep learning methods have exhibited remarkable advancements, attaining notable levels of performance in depression detection. \cite{cong2018xa} proposed a deep learning model named X-A-BiLSTM for depression detection in imbalanced social network data. Leveraging the capabilities of the Transformer model, \cite{malviya2021transformers} demonstrated significantly improved accuracy in detecting depression among social network users. 

Nevertheless, the existing methods predominantly focus on using pre-trained models or deeper networks to obtain the semantic aspects of sentences. The neglect of sentiment features of the target sentences and external sentiment knowledge leads to unsatisfactory performance~\cite{chancellor2020methods}\cite{tsugawa2015recognizing} of the neural network in depression detection. Because depression is usually highly correlated with negative emotions, these sentiment features could contribute to more comprehensive and accurate depression detection. With the intuition that sentiment serves as a direct clue to depression~\cite{park2012depressive}, we introduce external sentiment knowledge into depression detection to enhance performance. Specifically, our approach incorporates external sentiment knowledge into the depression detection model by leveraging a multi-task learning framework. This framework facilitates the simultaneous learning of sentiment analysis and depression detection, enabling the model to benefit from the additional information provided by external sentiment knowledge. The main contributions of this work are summarised as follows:

\begin{enumerate}
    \item We propose a Depression detection based on Sentiment Knowledge model called \textbf{DeSK}, which employs multi-task learning to acquire and leverage external sentiment knowledge which is overlooked by previous works.
    
    \item Considering the scarcity of publicly available datasets pertaining to depression in Chinese social networks, we collect and construct a dataset focused on depression from the Weibo platform. This dataset was created through self-diagnosis methods for further research and analysis\footnote{Contact the second author to inquire about the dataset.}. 
    
    \item Experimental results on the Reddit dataset show that DeSK outperforms state-of-the-art performance. The ablation tests validate the model components and demonstrate their efficacy in detecting depression.
\end{enumerate}

\section{Related Work}

Depression detection and sentiment analysis have been extensively studied in the field of mental health analysis using social network data. Early approaches focused on utilizing statistical or traditional machine learning methods to detect depression based on textual content. These methods mainly relied on linguistic patterns and semantic information to identify signs of depression. However, they often overlooked the crucial role of sentiment features in depression detection. However, multi-task learning offers a promising solution to address this disadvantage. By jointly learning these tasks, the models can benefit from shared knowledge and improve performance in both domains.
\subsection{Depression Detection}
Trained professional psychologists rely on various methods, including written descriptions provided by individuals and psychometric assessments, to assess and diagnose depression accurately~\cite{orabi2018deep}. Social network-based sentiment analysis is an alternative depression detection approach rising in recent years. Researchers can extract valuable insights from the vast amount of data available on social networks, such as patterns, trends, and user-generated content related to depression. 

By extracting various behavioral attributes from social network platforms, such as social engagement, mood, speech and language style, self-networking, and mentions of antidepressants, \cite{de2013predicting} aimed to provide estimates of depression risk. \cite{cong2018xa} proposed a deep learning model (X-A-BiLSTM) to handle the real-world imbalanced data distributions in social networks for depression detection. \cite{lin2020sensemood} proposed a deep visual-textual multi-modal learning approach aimed at acquiring robust features from both normal users and users diagnosed with depression. And \cite{guo2023leveraging} developed a depression lexicon based on domain knowledge of depression to facilitate better extraction of lexical features related to depression. 

\subsection{Sentiment Analysis}
Sentiment analysis seeks to examine individuals' sentiments or opinions concerning various entities, including but not limited to topics, events, individuals, issues, services, products, organizations, and their associated attributes~\cite{yue2019survey}. For the past few years, the growth of social networks has significantly propelled the advancement of sentiment analysis. To date, the majority of sentiment analysis research is based on natural language processing techniques. \cite{gaind2019emotion} combined text features and machine learning methods to classify social network texts into six types of emotions. \cite{suhasini2020emotion} employed machine learning algorithms, specifically Naive Bayes (NB) and the k-nearest neighbor algorithm (KNN), to discern the emotional content of Twitter messages and classified the Twitter messages into four distinct emotional categories.

\subsection{Multi-task Learning}
Multi-Task Learning (MTL) is a machine learning paradigm that aims to enhance the generalization performance of multiple related tasks by leveraging the valuable information inherent in these tasks~\cite{zhang2021survey}. \cite{zhang2014facial} proposed a facial landmark detection by combining head pose estimation and facial attribute inference. \cite{zhou2021hate} incorporated sentiment knowledge into a hate speech detection task by employing a multi-task learning framework. 

\begin{figure}[!t]
\centering
\includegraphics[scale=0.25]{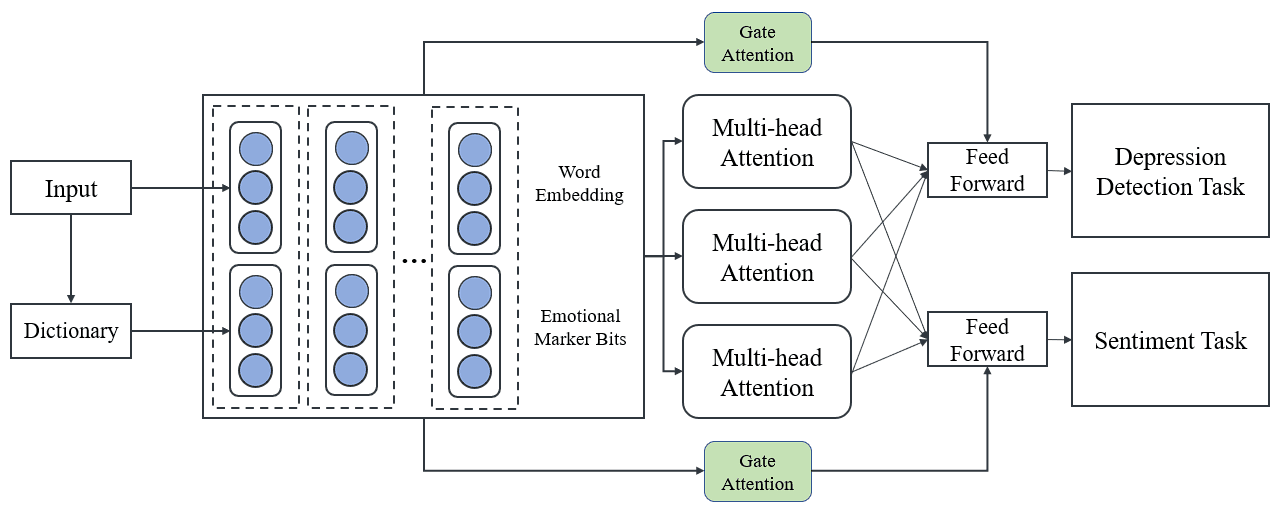}
\caption{The overall framework of DeSK.}
\label{fig:1}
\end{figure}

\section{Methodology}
In this section, we introduce DeSK, which exhibits an enhanced capability to detect depression through the integration of target sentence sentiment and external sentiment knowledge. The overall architecture is shown in Figure~\ref{fig:1}. The framework primarily comprises three components: 1) Input layer, which captures the sentiment features of the sentence. A depressed words dictionary is employed to determine if each word exhibits depressed speech characteristics and appended to the word embedding as emotional marker bits. 2) Multi-task learning framework, which leverages the strong correlation between sentiment analysis and depression detection, to model task relationships and acquire task-specific features by leveraging shared sentiment knowledge. Multiple feature extraction units, consisting of a multi-head attention layer and a feed-forward neural network, are utilized for this purpose. 3) Gated attention layer, which is a gated attention mechanism that calculates the probability of selecting each feature extraction unit. 

\subsection{Input Layer}
The central idea of DeSK revolves around the notable association between depression and negative emotions. We hypothesize that texts expressing depressive sentiments frequently contain explicit usage of negative emotion words~\cite{tolboll2019linguistic}. Hence, directing attention towards capturing derogatory words within a sentence can assist in enhancing depression detection capabilities. More specifically, we capture this sentiment information by utilizing sentiment marker bits.

\textbf{Sentiment Marker Bits.} Our work is based on the intuition that certain specific words that possess an exceedingly negative nature, such as sadness~\cite{mouchet2022sadness}, disgust~\cite{ypsilanti2019self}, etc., have a more substantial impact on the assessment of depression. To address this, we have constructed a depressed word dictionary, the vocabulary of which comes from NRC Emotion-Lexicon~\cite{mohammad2013nrc}. The depressed word dictionary is employed to classify social network text into two categories: those containing depressed words and those without depressed words. Each word in the text is assigned to one of these categories. The category assignment for each word is initialized randomly as a vector which we call sentiment marker bits: $ S = (s_1, s_2, s_n)$.

\textbf{Word Embedding.} Our word embedding is based on distributed representations of Words~\cite{mikolov2013distributed}, mapping words into a high-dimensional feature space while  preserving their semantic information. For each text, we represent it as $T = \{w_1, w_2, w_n \}$ using word embedding, where $w_i\in\mathcal{R^D}$ denotes each token embedding and $\mathcal{D}$ is dimensions of word vectors. 

Due to the linear structure observed in typical word embedding representations, it becomes feasible to meaningfully combine words by element-wise addition of their vector representations. To effectively leverage the information contained within depressed words, we integrate each word embedding with the sentiment marker bits. Regarding the implementation aspect, a simple vector concatenation operation is used, and the embedding of a word $v_i$ is calculated as $v_i = w_1 \bigoplus s_1$.

\subsection{Multi-task Learning Framework}
Considering the diverse influences of various countries, regions, religions, and cultures, some meanings in many languages are often embedded within the underlying semantics rather than solely reflected in sentiment words. For instance, the word ``blue'' may not explicitly convey a sense of depression, but it often carries a pessimistic semantic meaning. As evident from the aforementioned example, depression text frequently includes negative sentiment words. However, relying solely on the sentiment information within the target sentence itself for depression detection often proves challenging in achieving satisfactory performance. 

The task of determining the sentiment of a text based on its semantic information is commonly referred to as sentiment analysis. Extensive research has been conducted on sentiment analysis for many years, resulting in the availability of abundant high-quality labeled datasets. In contrast, in the depression detection field, the availability of high-quality labeled data is limited, leading to a restricted vocabulary and inherent biases during the training process. In multi-task learning, the commonly used framework employs a shared-bottom structure where different tasks share the bottom hidden layer. While this structure can mitigate overfitting risks, its effectiveness may be impacted by task dissimilarities and data distribution. In DeSK, we incorporate multiple identical feature extraction units that share the output from the previous layer as input and pass it on to the subsequent layer. This allows for an end-to-end training of the entire model. Our feature extraction units layer consists of a multi-head attention layer and two feed-forward neural networks.

\textbf{Multi-head Attention Layer.} To capture long-distance dependencies between words in a sentence, we employ the multi-head self-attention mechanism introduced by \cite{vaswani2017attention}. This approach calculates the semantic similarity and semantic features of each word in the sentence with respect to other words, allowing for enhanced connectivity and information exchange throughout the sentence. The formula is as follows:

For a given query $Q\in\mathcal{R}^{n_1 \times d_1}$, key $K\in\mathcal{R}^{n_1 \times d_1}$, value $V\in\mathcal{R}^{n_1 \times d_1}$,$$Attention(Q,K,V) = softmax(\frac{QK^T}{d_1})V$$

For the i-th attention head, let the parameter matrix $W^Q_i$, $W^K_i$, $W^V_i$, $$M_i = Attention(QW^Q_I, KW^K_I, VW^V_i)$$ 

The final feature representation is as follows:
$$H^s = concat(M_1, M_2,...,M_l)W_o$$

\textbf{Pooling Layer.} Based on the observation~\cite{shen2018use} that using a combination of max-pooling and average-pooling yields significantly better performance compared to using a single pooling strategy alone. By leveraging both pooling strategies, we are able to capture different aspects and variations within the features, leading to improved overall performance. The formula is as follows: 
$$P_m = Pooling_max(H^s)$$
$$P_a = Pooling_average(H^s)$$
$$P_s = concat(P_m, P_a)$$
\subsection{Gated Attention Layer}
The gated attention mechanism enables the model to dynamically select a subset of feature extraction units based on the input. Each task has its own gate, and the weight selection varies for different tasks. The output of a specific gate represents the probability of selecting a different feature extraction unit. Multiple units are then weighted and summed to obtain the final representation of the sentence, incorporating the contributions from the selected feature extraction units.

 The formula is as follows:
 $$g^k(x) = softmax(W_{gn} * gate(x))$$
 $$f^k(x) = \sum^{n}_{i=1}g^k(x)_if_i(x)$$
 $$y_k = h^kf^k(x)$$
 where k is the number of tasks.

 \subsection{Model Training}
For the training process, the loss function used in DeSK combines cross entropy with L2 regularization, as follows:
$$loss = -\sum\limits_i\sum\limits_j y_i^j\log{\hat{y}_i^j}+\lambda\|\theta\|^2$$
where $i$ is the index of sentences, and $j$ is the index of class. 

\section{Experiment}
In this section, we begin by introducing the datasets used in our study as well as the evaluation metrics employed for performance assessment. Then, we present a series of ablation experiments conducted to showcase the effectiveness of DeSK. Finally, we provide a comprehensive analysis of the results obtained from these experiments.

\subsection{Datasets}
To explore whether sharing sentiment knowledge can enhance the performance of hate depression detection, we utilize three publicly available datasets for social network depression detection and one sentiment dataset. Meanwhile, to validate the cross-language nature of the model, we use another depression detection dataset which we constructed, and a sentiment analysis dataset in Chinese. The specifics of these datasets are presented in Table~\ref{tab:1}.

\textbf{Reddit Depression Dataset (RDD)} is collected~\cite{sampath2022data} from the archives of subreddit groups such as ``r/MentalHealth,'' ``r/depression,'' ``r/loneliness,'' ``r/stress,'' and ``r/anxiety.'' These subreddits provide online platforms where individuals share their experiences and discussions related to mental health issues. The collected postings data were annotated by two domain experts who assigned three labels to denote the level of signs of depression: ``Not depressed,'' ``Moderate,'' and ``Severe.'' 

\textbf{Tweet Depression Dataset 60k (TDD-60k)} utilized in our study consists of four depression labels~\cite{tavchioski2023detection}, each corresponding to different levels of depression signs. By incorporating these four depression labels, it aims to capture a comprehensive range of depression severity in the Twitter dataset.

\textbf{Tweet Depression Dataset 10k (TDD-8k)} is from huggingface\footnote{https://huggingface.co/datasets/ShreyaR/DepressionDetection}. The dataset exhibits a near-equivalent distribution of positive and negative examples, indicating a balanced representation between the two classes.

\textbf{Sentiment Analysis (SA)}  is  from Kaggle2018\footnote{https://www.kaggle.com/dv1453/twitter-sentiment-
analysis-analytics-vidya}. The SA dataset exhibits a higher number of positive cases but a relatively smaller number of negative cases. As the test set does not have labeled data, we solely rely on the training set for our analysis and model training.

\textbf{Chinese Weibo Depression Dataset (CWDD)} was compiled by collecting tweets from depression-related communities on Weibo. It underwent annotation and organization to ensure a balanced distribution of positive and negative cases, thereby achieving comparable proportions between the two classes. The dataset consists of 10,348 tweets from Weibo classified as depressed, while there are 7,562 tweets categorized as non-depressed.

\textbf{Weibo Sentiment 100k (WS)} consists of over 100,000 comments from Sina Weibo, which have been tagged with emotion labels. It contains  59,993 positive comments and 59,995 negative comments, making it a balanced dataset in terms of positive and negative sentiment. In consideration of the proportion of sentiment data and depression data, we have specifically chosen 10,500 instances from the dataset for the purpose of training. This selection process ensures a balanced representation of both sentiment and depression-related data in the training set. 

Given the cross-language nature of our experiments, it is essential to construct separate depression word dictionaries for English and Chinese. The English depression dictionary is constructed with reference to the NRC Emotion Lexicon~\cite{mohammad2013nrc}. The NRC Emotion Lexicon is a widely recognized resource that provides a comprehensive collection of words annotated with their associated emotions. 
Our Chinese depression word dictionary is constructed based on the Dalian University of Technology Chinese Emotion Vocabulary Ontology Database~\cite{xu2008constructing}. This database is a recognized and comprehensive resource that contains a wide range of Chinese words and phrases annotated with their corresponding emotional categories.

\begin{table}
    \caption{Statistics of datasets used in the experiment.}
    \setlength{\tabcolsep}{9mm}
    \label{tab:1}
    \begin{tabular}{ccc}
    \toprule
        Dataset & total & Classes \\
    \midrule
        RDD & 156,676 &
        \makecell{Not depressed (4,649)\\ Moderately depressed (10,494)\\Severely depressed (1,489)} \\[10pt]
        TDD-60k& 66,228 & \makecell{Not depressed (27,202)\\ Moderately depressed(32,121)\\Severely depressed (9,968)} \\[10pt]
        TDD-7k & 7,771 & \makecell{Not Depressed (3,900)\\ Depressed (3,832)} \\
        CWDD & 17,910 & \makecell{Not Depressed (10,348)\\ Depressed (7,562)} \\
        \hline
        SA & 31,962 & \makecell{negative (2,242)\\ positive (29,720)} \\
        WS & 119,988 & \makecell{negative (59,995)\\ positive (59,993)} \\
    \bottomrule
    \end{tabular}\par\smallskip
\end{table}

\subsection{Baselines and Metrics}

\noindent\textbf{Baselines.} We compare the performance of DeSK with the following baselines to evaluate its effectiveness. 

\textbf{Doc2vec} is proposed by~\cite{mikolov2013distributed}, known as an unsupervised learning algorithm that aims to represent documents as fixed-length numerical vectors. It is an extension of the popular Word2Vec algorithm, which is used to generate word embeddings.

\textbf{BERT} is proposed by~\cite{devlin2018bert}. The pre-trained model BERT was used to capture the features of depression detection.  

\textbf{RoBERTa}, proposed by~\cite{liu2019roberta}, is an enhanced version of the BERT model, which introduces various optimizations to improve its performance. These optimizations focus on refining the underlying architecture and training process.\\

\noindent\textbf{Metrics.} In the depression detection task, we employ two evaluation metrics, namely Accuracy (ACC) and Macro F1, to assess the performance of DeSK.

\subsection{Training Details}
In the experiments, we use the following configurations for different components of the model. In the input layer, we initialize all word vectors using Glove Common Crawl Embeddings (840B Token) with a dimension of 300. The category embeddings, on the other hand, are initialized randomly with a dimension of 100.

For the sentiment knowledge sharing layer, we employ a multi-head attention mechanism with four heads. The first Feed-Forward network consists of a single layer with 400 neurons, while the second Feed-Forward network includes two layers with 200 neurons each. Dropout is applied after each layer with a dropout rate of 0.1.

The RMSprop optimizer is utilized with a learning rate of 0.001, and the models are trained using mini-batches consisting of 512 instances. To prevent overfitting, we incorporate learning rate decay and early stopping techniques during the training process. These measures ensure effective model training and help mitigate the risk of overfitting.

\subsection{Model Performance}
\begin{table}[ht]
    \caption{Comparison of DeSK performance with the baseline in RDD.}
    \setlength{\tabcolsep}{13mm}
    \label{tab:2}
    \begin{tabular}{ccc}
    \toprule
        Model & Accuracy & Macro F1 \\ 
    \midrule
        Doc2vec & 54.1 & 54.9 \\[10pt]
        BERT & 55.9 & 56.1 \\[10pt]
        RoBERTa & 55.7 & 56.3 \\[10pt]
        \textbf{DeSK} & \textbf{62.7} & \textbf{60.8} \\[10pt]        
    \bottomrule
    \end{tabular}\par\smallskip
\end{table}

The overall performance comparison is summarized in Table~\ref{tab:2}. DeSK outperforms other neural network models in terms of accuracy and F1 score. Specifically, compared to Doc2Vec, DeSK achieves a 10\% increase in the F1 score. Even when compared to the strong baseline model, universal encoder, DeSK demonstrates superior performance. Furthermore, DeSK has the advantage of being easier to implement and having fewer parameters compared to other models. This makes it more accessible and efficient for practical applications.
\begin{table}[ht]
    \caption{Performance of DeSK on different datasets.}
    \setlength{\tabcolsep}{13.5mm}
    \label{tab:3}
    \begin{tabular}{ccc}
    \toprule
        Dataset & Accuracy & Macro F1 \\ 
    \midrule
        RDD & 62.7 & 60.8 \\[10pt]
        TDD-60k& 81.9 & 80.9 \\[10pt]
        TDD-7k & 89.6 & 89.1 \\[10pt]
        CWDD & 96.3 & 96.3 \\[10pt]
    \bottomrule
    \end{tabular}\par\smallskip
\end{table}
Additionally, we conducted tests on different datasets, including the Chinese Weibo dataset, to evaluate the cross-language capability of DeSK. The results are shown in Table~\ref{tab:3}, indicating that DeSK performs well in analyzing and processing text from different languages, showcasing its ability to handle multilingual data effectively. This versatility further enhances the applicability of DeSK across various linguistic contexts.

\subsection{Ablation Experiments}
\begin{table}[ht]
    \caption{The results of ablation experiments on TDD-60k dataset.}
    \setlength{\tabcolsep}{13.5mm}
    \label{tab:4}
    \begin{tabular}{ccc}
    \toprule
        Model & Accuracy & Macro F1 \\ 
    \midrule
        -ss & 77.4 & 77.3 \\[10pt]
        -s & 80.3 & 80.2 \\[10pt]
        DeSK & \textbf{81.9} & \textbf{80.9} \\[10pt]
    \bottomrule
    \end{tabular}\par\smallskip
\end{table}
We conducted an analysis to assess the impact of different components of DeSK. The results are presented in Table~\ref{tab:4}, where ``-ss'' refers to the ablation of sentiment knowledge sharing and sentiment marker bits, while ``-s'' indicates that sentiment data was not utilized as input and only sentiment marker bits were used in the model. The findings provide insights into the contributions of these components to the overall performance of the model.

\begin{table}[ht]
    \caption{The influence of gate mechanism on TDD-7k dataset.}
    \setlength{\tabcolsep}{13.5mm}
    \label{tab:5}
    \begin{tabular}{ccc}
    \toprule
        Model & Accuracy & Macro F1 \\ 
    \midrule
        -gate & 89.2 & 88.8 \\[10pt]
        DeSK & \textbf{89.6} & \textbf{89.1} \\[10pt]
    \bottomrule
    \end{tabular}\par\smallskip
\end{table}

We conducted an analysis to evaluate the impact of gated attention in DeSK. The results, as shown in Table~\ref{tab:5}, demonstrate that the model's performance is further improved when gated attention is utilized. By learning how the outputs of different gates interact and contribute to the final representation, DeSK can better understand the dependencies and correlations between tasks, leading to improved performance in capturing complex task relationships.

\begin{figure}[!t]
\centering
\includegraphics[scale=0.5]{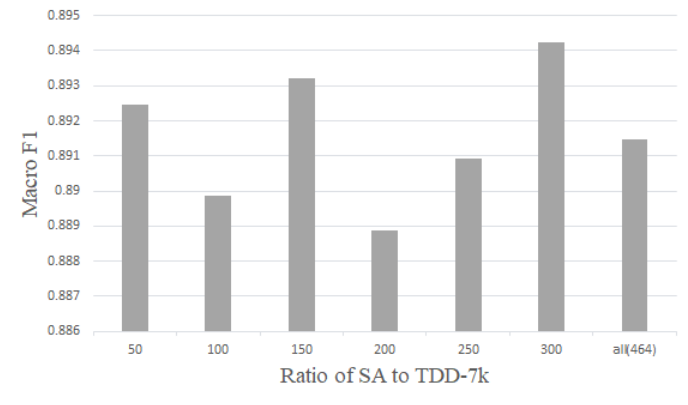}
\caption{The influence of the scale of sentiment data
set.}
\label{fig:2}
\end{figure}

\subsection{The Influence of Sentiment Dataset}
As mentioned earlier, depression detection and sentiment analysis exhibit a strong correlation, indicating that sharing sentiment knowledge can enhance the performance of depression detection. However, the relative data proportions of the two tasks in multi-task learning can influence the performance of each task~\cite{liu2019multi}. To investigate the influence of the dataset ratio on the performance of multi-task learning, we select the smallest dataset, TDD-7K, for the experiment. We then analyze the effect by sampling and adjusting the size of the SA (Sentiment Analysis) dataset.

Figure~\ref{fig:2} illustrates the performance of the model in relation to the ratio of data in the sentiment dataset and the depression dataset. It reveals that the model's performance tends to be poor when the ratio of data in the sentiment dataset is small compared to the depression dataset. Conversely, the model achieves its best performance when the ratio reaches 3:1. This finding suggests that having an appropriate proportion of data from both the sentiment dataset and the depression dataset is crucial for achieving optimal performance in multi-task learning. 

\section{Discussion}
Depression has emerged as a significant global health concern, affecting individuals across various geographical locations. With the widespread adoption of social networks and the optional anonymity they provide, many individuals, whether diagnosed or not, may express their mood or symptoms related to depression on these platforms. This presents an opportunity to identify individuals at risk of depression through their online activities. Detecting individuals at risk of depression on social networks has great potential. It allows for timely intervention and support for those who may struggle to access social support or effective treatment through traditional means. By leveraging the power of technology and analyzing online behavior, we have the potential to reach individuals who might otherwise go unnoticed and provide them with the necessary assistance they need.

\section{Conclusion and Future Work}
This paper focuses on investigating the effectiveness of multi-task learning in depression detection tasks. The core concept revolves around utilizing multiple feature extraction units to share multi-task parameters, thereby enabling improved sharing of sentiment knowledge. The proposed model incorporates gated attention to fuse features for depression detection. By leveraging both the sentiment information from the target and external sentiment resources, DeSK demonstrates enhanced system performance through ablation experiments, thereby advancing social network depression detection. Through detailed analysis, we provide further evidence of the validity and interpretability of DeSK.

Overall, our experiments contribute to a better understanding of the interplay between depression detection and sentiment analysis through multi-task learning. They lay the foundation for future endeavors in refining modeling techniques and data selection, encompassing different types of social network depression information, diverse sentiment data types and scales, and other related aspects. But there are still some limitations. One key limitation is the quality of the available depression detection datasets, which are often labeled based on self-diagnosis. This introduces potential biases and uncertainties in the data, which can impact the performance of depression detection. Additionally, while psychologists may have access to a wealth of depression-related information through counseling sessions, privacy concerns prevent the correlation of this information with social network data. This limits our ability to create a comprehensive and high-quality depression detection dataset that combines both social network information and professional insights. 

To overcome these limitations, we plan to explore privacy computing techniques to achieve alignment between social network data and individual entities without compromising privacy. This would allow us to create a more robust and reliable depression detection dataset, enhancing the accuracy and effectiveness of depression detection.

\section{Acknowledgments}


\end{document}